\documentclass[letterpaper]{article} % DO NOT CHANGE THIS
\usepackage{aaai25}  % DO NOT CHANGE THIS
\usepackage{times}  % DO NOT CHANGE THIS
\usepackage{helvet}  % DO NOT CHANGE THIS
\usepackage{courier}  % DO NOT CHANGE THIS
\usepackage[hyphens]{url}  % DO NOT CHANGE THIS
\usepackage{graphicx} % DO NOT CHANGE THIS
\usepackage{natbib}  % DO NOT CHANGE THIS AND DO NOT ADD ANY OPTIONS TO IT
\usepackage{caption} % DO NOT CHANGE THIS AND DO NOT ADD ANY OPTIONS TO IT
\usepackage{algorithm}
\usepackage{algorithmic}
\usepackage{newfloat}
\usepackage{listings}
\DeclareCaptionStyle{ruled}{labelfont=normalfont,labelsep=colon,strut=off} % DO NOT CHANGE THIS
\floatstyle{ruled}
\newfloat{listing}{tb}{lst}{}
\floatname{listing}{Listing}

\usepackage{amssymb}
\usepackage{multirow}
\usepackage{arydshln}  %
\usepackage{booktabs}
\usepackage{float}
\usepackage{subfigure}
\usepackage{color}
\usepackage{xcolor}
\usepackage{url}
\usepackage{marvosym}

\title{ParGo: Bridging Vision-Language with Partial and Global Views}
\author {
    \normalsize
    An-Lan Wang\textsuperscript{\rm 1, 2, $\dagger$,}\thanks{ This work is done when An-Lan Wang is an intern at ByteDance China.},
    Bin Shan\textsuperscript{\rm 2, }\thanks{Equal contribution. \\ \indent\ \textsuperscript{\Letter}Corresponding authors.},
    Wei Shi\textsuperscript{\rm 2},
    Kun-Yu Lin\textsuperscript{\rm 1},
    Xiang Fei\textsuperscript{\rm 2}, 
    Guozhi Tang\textsuperscript{\rm 2}, 
    Lei Liao\textsuperscript{\rm 2}, \\
    Can Huang\textsuperscript{\rm 2}, 
    Jingqun Tang\textsuperscript{\rm 2, \Letter}, 
    Wei-Shi Zheng\textsuperscript{\rm 1, \Letter}
}
\affiliations {
    \textsuperscript{\rm 1}School of Computer Science and Engineering, Sun Yat-sen University, China\\
    \textsuperscript{\rm 2}ByteDance China\\
    \{wanganlan, shanbin, shiwei.11, feixiang.77, tangguozhi.997, liaolei.666, can.huang tangjingqun\}@bytedance.com, \\ linky5@mail2.sysu.edu.cn, wszheng@ieee.org
}
\usepackage{bibentry}
\begin{document}
\maketitle
\begin{abstract}
This work presents ParGo, a novel Partial-Global projector designed to connect the vision and language modalities for Multimodal Large Language Models (MLLMs). 
Unlike previous works that rely on global attention-based projectors, our ParGo bridges the representation gap between the separately pre-trained vision encoders and the LLMs by integrating global and partial views, which alleviates the overemphasis on prominent regions. 
To facilitate the effective training of ParGo, we collect a large-scale detail-captioned image-text dataset named ParGoCap-1M-PT, consisting of 1 million images paired with high-quality captions. 
Extensive experiments on several MLLM benchmarks demonstrate the effectiveness of our ParGo, highlighting its superiority in aligning vision and language modalities. 
Compared to conventional Q-Former projector, our ParGo achieves an improvement of 259.96 in MME benchmark. 
Furthermore, our experiments reveal that ParGo significantly outperforms other projectors, particularly in tasks that emphasize detail perception ability. 
\end{abstract}
\begin{links}
    \link{Code and models}{https://github.com/bytedance/ParGo}
\end{links}

\section{Introduction}
Recent Multi-Modal Large Language Models (MLLMs) \cite{Achiam2023GPT4TR,team2023gemini,liu2023llava,li2022blip} achieve remarkable progress across various tasks (\textit{e.g.}, Visual Question Answering). 
The vision-language projector as a widely used component in MLLMs, aims to provide LLMs with proper visual features. Due to its critical role in bridging modalities, it has garnered significant attention in recent research \cite{cha2023honeybee,alayrac2022flamingo,zhu2023minigpt4}.

\begin{figure}[t]
  \includegraphics[width=\columnwidth]{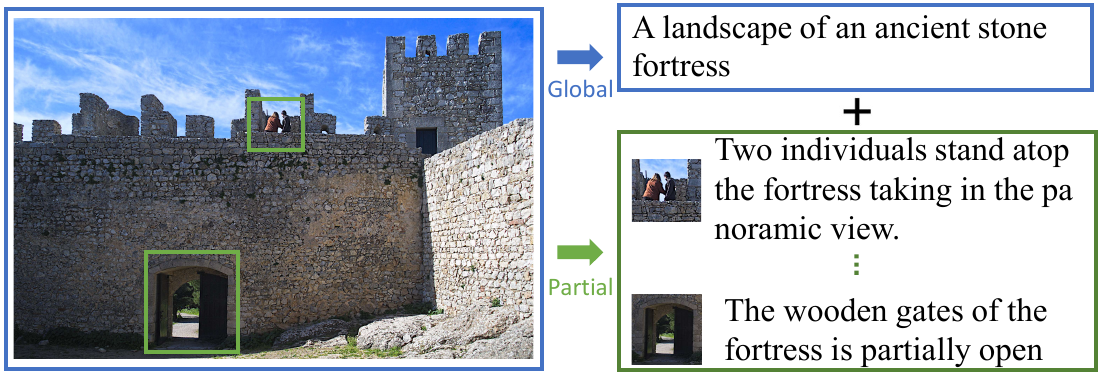}
  \caption{Illustration of the global and partial information. 
  An image can be properly described by the two kinds of information.
  Globally, this image shows a landscape of an ancient stone fortress. 
  Delve into the partial information, two individuals stand atop the fortress, the wooden gate at the bottom of the fortress is partially open, and so forth. 
  }
  \label{fig:global_partial}
\end{figure}

The pioneer works \cite{zhu2023minigpt4,liu2023llava} directly project the visual feature using linear or Multi-Layer Perceptron layer (MLP). 
Nevertheless, such linear-based projector struggles to control the number of visual tokens provided to LLMs (\textit{e.g.}, handling fine-grained features), resulting in high computational costs. 
Another line of works \cite{li2023blip2, alayrac2022flamingo}, employing global attention-based projectors, perform a global projection of image features to a fixed number of visual tokens using attention operation. 
However, these projectors based on global projection lead to the produced tokens concentrating on prominent regions while overlooking finer details. 
Take the image in Figure~\ref{fig:global_partial} as an example, previous methods tend to focus on the fortress and easily overlook the two individuals at the top.

In this paper, we aim to build a vision-language projector that can provide the LLMs with visual features that better represent the image, while using a fixed number of visual tokens. 
The inspiration stems from the observation that an image can be properly described by two kinds of information, namely, global information presents a holistic understanding of images, while multiple partial information emphasizes the subtle details, an example is shown in Figure \ref{fig:global_partial}. 

Motivated by this, we propose a novel Partial-Global projector (ParGo) based on a partial-global attention mechanism. 
By integrating both global and partial views, our ParGo effectively bridges the representation gap between separately pre-trained vision encoders and LLMs, alleviating the overemphasis on prominent regions. 
In addition, considering the relation between different partial regions in an image, ParGo incorporates a cascaded partial perception block, which enables interaction between different partial regions of an image. 

Furthermore, to facilitate the effective training of ParGo, we collect a large-scale detail-captioned image-text dataset named ParGoCap-1M-PT for pre-training. 
Most existing pre-training datasets, typically sourced from the Internet, contain captions that are usually short and emphasize prominent visual features while lacking detailed descriptions of partial regions. 
Training on such datasets makes it challenging for the model to learn fine-grained details. 
In contrast, our ParGoCap-1M-PT contains longer and more detailed descriptions of multiple regions in images.
Pre-trained on the two kinds of captioned data, we transfer our models into multiple downstream tasks using several public-available instruction tuning datasets, \textit{e.g.}, LLaVA-150k~\cite{liu2023llava1.5}. 
Extensive experiments are conducted on several benchmarks, and the results demonstrate that our Partial-Global projector outperforms other projectors. 
Our contribution can be summarized:
\begin{itemize}
\item We propose a novel Partial-Global projector (ParGo) that well aligns two separately pre-trained models by integrating partial and global views, alleviating the overemphasis on prominent regions. 
\item To facilitate the modality alignment, we further propose a new detail-captioned pre-training dataset, ParGoCap-1M-PT, including 1 million images paired with high-quality captions.
\item Extensive experiments on several MLLM benchmarks demonstrate the effectiveness of our proposed ParGo. 
Our projector achieves state-of-the-art results compared with previous projectors, particularly excelling in some tasks that need more detail-perception ability. 
\end{itemize}

\section{Related Work}
\subsection{Multi-modal Large Language Models}
Large Language Models (LLMs) \cite{ touvron2023llama, Achiam2023GPT4TR, chiang2023vicuna} have achieved remarkable progress, leading recent works \cite{alayrac2022flamingo,team2023gemini,bai2023qwen-vl} to generalize this success to more modalities, \textit{i.e.,}, Multimodal Large Language Models (MLLMs). 
In those works, closed-source works \cite{Achiam2023GPT4TR,team2023gemini,bai2023qwen-vl} have shown great advancements, highlighting their high performance on complex tasks. In contrast, open-source models have also made significant progress, promoting transparency and collaboration in the research community. 
Pioneer works \cite{li2023blip2,alayrac2022flamingo, shan2022ernie, shan2022ernie2} established competitive baselines by integrating massive image-text pairs. 
Furthermore, recent works \cite{liu2023llava,dai2024instructblip} \cite{liu2023llava1.5, liu2023llava,chen2023sharegpt4v} boosting the zero-shot capabilities on various downstream tasks via collecting more high-quality multi-modal instruction. 
On the other hand, recent works \cite{liu2024llavanext,hu2023bliva,liu2024textmonkey,zhao2024multi,zhao2024harmonizing,liu2023spts, li2024techcoach, lin2024rethinking} also focus on fine-grained understanding (\textit{e.g.}, text recognition, action understanding).

\begin{figure*}[t]
  \includegraphics[width=1\linewidth]{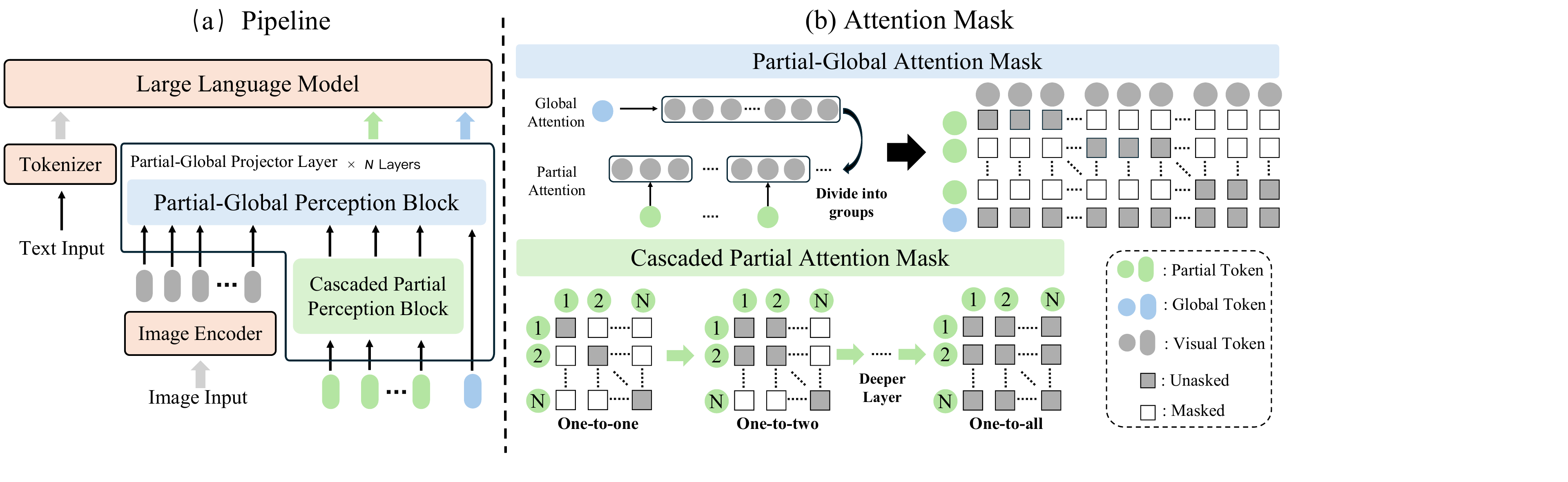}
  \caption{(a). The pipeline of a MLLM with our proposed ParGo as the vision-language projector. 
  First of all, we use a frozen image encoder to extract image features. 
  To better align the pre-trained visual encoder with the LLM, we propose a Partial-Global projector to project the image features using two kinds of tokens \textit{i.e.,} partial and global tokens. 
  Finally, the output partial and global visual tokens, as well as the tokenized text, are fed into the LLM to generate the text output in an auto-regressive manner. 
  Specifically, each Partial-Global projector layer contains a Partial-Global Perception block that utilizes two kinds of tokens to extract the image features. 
  Additionally, to fully consider the relation between different partial regions in an image, a cascaded partial perception block is incorporated to enable interactions between partial tokens in a cascaded manner. 
  (b). A Demonstration of the Partial-Global and the Cascaded Partial Attention mask. 
  It's worth noting that the Partial-Global Attention mask remains the same in different layers, while the Cascaded Partial Attention mask changes across various layers. 
  }
  \label{fig:pipeline}
\end{figure*}
\subsection{Vision-language Projector}
Vision-language projectors play a crucial role and are widely used components in MLLM. They aim to connect the visual feature space and language feature space, which can be divided into linear-based and attention-based projectors. 
Linear-based projectors \cite{liu2023llava,liu2023llava1.5,zhu2023minigpt4,chen2023sharegpt4v,internlmxcomposer2} employ a linear layer to connect the vision encoder seamlessly with the language model (LLM). 
Despite their straightforward implementation, the linear-based projectors encounter challenges in producing a large number of visual tokens to LLMs, leading to high computational costs. 
Another line of research~\cite{alayrac2022flamingo,li2023blip2,bai2023qwen-vl,dai2024instructblip,ye2023mplugowl2} explore more flexible projectors (\textit{e.g.}, Q-former \cite{li2022blip} and Perceiver Resampler \cite{alayrac2022flamingo}) based on attention mechanism. 
Such attention-based methods often extract prominent image features, leading to a loss of detail and a drop in the model performance. 
Similar findings are also mentioned in a recent work, Honeybee \cite{cha2023honeybee}, which proposes a D-Abstractor that uses a Deformable attention \cite{zhu2020deformableattention} to retain the local information and achieve superior performance. 
To efficiently provide comprehensive information to LLMs using a fixed number of visual tokens, we propose Partial-Global projector, which uses a partial-global projection that simultaneously extracts both partial and global information. 

\subsection{Multi-modal Pre-training Data}
Training models using web-crawled large-scale image-text datasets (\textit{e.g.}, \cite{schuhmann2021laion400m,kakaobrain2022coyo-700m,changpinyo2021cc12m,sharma2018conceptual}) has become the most common strategy for MLLMs. 
Nevertheless, web-crawled datasets primarily present the main feature of the image using noisy and short captions, lacking detailed descriptions. 
To obtain detailed descriptions, some works \cite{wang2023all,wang2025all} provide boxes (or mask) level captions but are constrained by the box-generation (grounding) model \cite{fang2024your, fang2025rethinking, tang2022optimal,tang2022few}.
The recent remarkable progress achieved by close-sourced MLLMs has led recent researchers \cite{chen2023sharegpt4v,yu2024capsfusion} to consider using MLLM to synthesize detail-captioned data, supplementing the limitations of conventional web-crawled datasets. 
In this work, we further contribute a detail-captioned dataset for pre-training, aimed at enhancing the alignment between the two modalities from a data perspective.

\section{Methodology}
\subsection{Overview}
In Figure~\ref{fig:pipeline}. (a), we illustrate the pipeline of the MLLM that uses our proposed Partial-Global projector (ParGo) as the vision-language projector. 
Given an image $I$ and related text $T$, we first use a frozen image encoder and a tokenizer to extract the visual feature $f_v$ and text feature $f_t$, respectively. 
To effectively align the vision and language modalities, we propose a Partial-Global projector to project the visual feature to the text feature space. 
Specifically, the Partial-Global projector projects the visual features from the partial and global views, employing two kinds of learnable tokens. 
Subsequently, the outputs of ParGo and the tokenized input text are fed into the Large Language Model to generate the final text output. 

\subsection{Partial-Global Projector}
To better align the separately pre-trained visual encoder and large language models, we propose the Partial-Global projector (ParGo). 
Each ParGo layer mainly contains two parts, \textit{i.e.}, a Partial-Global Perception block and a Cascaded Partial Perception block, as illustrated sequentially,.

\paragraph{Partial-Global Perception Block. }
Firstly, we propose a Partial-Global Perception block, which employs two kinds of tokens, \textit{i.e.,} partial tokens and global tokens, to extract the partial and global information, respectively. 

In detail, given the visual features $f_v \in \mathbb{R}^{n_v \times c}$ (extracted by the visual encoder), we randomly initialize a set number of global tokens $q_g \in \mathbb{R}^{n_g \times c}$ and partial tokens $q_p \in \mathbb{R}^{n_p \times c}$, where $c$ is the feature dimension, and $n_v$, $n_g$, $n_p$ is the number of visual features, global tokens, and partial tokens, respectively. 
The tokens interact with the image features in a cross-attention layer. 
For the cross-attention mask, we use a pre-defined mask $M\in \mathbb{R}^{(n_p+n_g) \times n_v}$, which ensures that one partial token only interacts with part of the visual features, while the global tokens interact with all visual features. 
The number of visual features each partial token interact with $n_s$ is calculated as follows:
\begin{equation}
    n_s = n_v / n_p,
\end{equation}
An example of the mask used in the Partial-Global Perception block is shown in Figure~\ref{fig:pipeline}.(b). 
Intuitively, each partial token only \textit{sees} part of the image and global tokens \textit{see} the whole image. 

\paragraph{Cascaded Partial Perception Block. }
In the above partial-global attention process, the partial tokens \textit{see} consistent visual tokens in different layers. 
However, such a process focuses on using one token to project part of the image, thus it may be unable to fully consider the relation between different partial regions in an image. 
Therefore, to fully consider the context of the partial image, we propose a Cascaded Partial Perception (CPP) block that enables interaction between different partial tokens. 

Specifically, in addition to the Partial-Global Perception block, we insert a Cascaded Partial Perception (CPP) block in front of it.
The CPP block is implemented using a masked self-attention block with a specially designed mask as the core.
The input is the partial tokens $\{q_{p}^i\}_{i=1}^{n_p}$, and as the depth $l$ increases, the number of adjacent tokens each partial token can \textit{see} $n_{vis}^{l} (0 \leq n_{vis}^l\leq n_{p})$ increase linearly as well. 
This process can be formulated as follows: 
\begin{equation}
n_{vis}^{l} = k \times l, 
k = n_{p} / d
\end{equation}
where $l$ is the index of the layer, $n_p$ is the number of partial tokens, and $d$ is the number of the Partial-Global projector layer. 
As shown in Figure~\ref{fig:pipeline}. (b), an example of the mask in different CPP blocks in different layers is visualized. 

\subsection{ParGoCap-1M-PT}
In this section, we focus on how to pre-train the overall model to leverage the advantage of the Partial-Global projector. 
Existing pre-training corpora are generally coarse-captioned, and collected from the Internet. 
Such captioned data are noisy and are usually short (average character number less than 100), describing only part of the image. 
Using such coarse data hinders the alignment between the two modalities. 
Therefore, we construct a new large-scale detail-captioned image-text dataset, named \textbf{ParGoCap-1M-PT}, using off-the-shell closed-source MLLMs. 
Table~\ref{tav:data_stat} compares existing caption datasets and our ParGoCap-1M-PT. 
ParGoCap-1M-PT offers a large amount of high-quality detailed captioned samples, which distinguishes itself from existing datasets.
The data collection pipeline mainly consists of two steps: 

\paragraph{Detailed caption generation. }
To facilitate the alignment between the vision and language feature space, large-scale and diversified image data is required. 
We first randomly select a large number of images from the Laion dataset~\cite{schuhmann2021laion400m}. 
Then, to generate detailed captions that well describe the image, we employ the powerful close-sourced MLLMs (\textit{i.e.,} GPT4-V and Gemini) to generate the captions given a specified prompt. 
Since our goal is to generate captioned data that takes into account both partial and global information about an image, we design the prompt that asks the MLLMs to describe the image globally and partially. 

\paragraph{Quality control. }
Thanks to the powerful capabilities of existing models \cite{chatgpt, team2023gemini}, the quality of the generated data is already very excellent. 
However, there may still be some erroneous data due to the hallucination problem. 
To further filter out high-quality data, we employed a simple but effective quality control method. 
Following previous works~\cite{li2022blip}, we directly use several models \cite{radford2021clip,li2022blip} to calculate the similarity between the image and generated captions. 
Image-caption pairs with low similarity are dropped. 
This step filters out a small portion of the data, proving that our data quality is exceptional. 
For more dataset details, please refer to the supplementary materials. 

\section{Experiment}
\subsection{Benchmarks and Metrics}
To thoroughly validate the superiority of the proposed ParGo, we utilize four benchmarks tailored specifically for evaluating Multi-modal Large Language Models (MLLMs), including MME \cite{fu2023MMEBench}, MMBench \cite{liu2023mmbench}, SEED-Bench~\cite{li2023seedbench}, and MM-Vet \cite{yu2024mmvet}. 
The First three benchmarks evaluate a range of MLLM capabilities, including perceptual understanding and visual reasoning. 
They employ different question formats: MME uses binary yes/no questions, while MMBench and SEED-Bench utilize multiple-choice questions.
As for the MM-Vet, which uses the GPT-4 to evaluate the open-ended outputs of MLLMs, we include this benchmark to monitor the model's performance in natural language generation. 

For all benchmarks, we report the official metrics computed using
official implementation.

\begin{table}[t!]
    \renewcommand\arraystretch{1.2}
    \centering
    \footnotesize
    \resizebox{1.0\linewidth}{!}{
    \setlength{\tabcolsep}{0.5mm}{
    \begin{tabular}{l|ccc}
    \hline
    Dataset  & Captioned by  & Samples & Avg.  \\ 
    \hline
    \multicolumn{4}{l}{\textit{\textbf{Coarse-captioned}}} \\
    \hline
    COCO-Caption \cite{chen2015microsoft}  & Human & 118K & 52 \\
    BLIP-LCS \cite{li2022blip}   & BLIP & 558K & 54 \\
    LLaVA-23K \cite{liu2023llava}  & GPT4 & 23K & 609 \\ 
    \hline
    \multicolumn{4}{l}{\textit{\textbf{Detail-captioned}}} \\
    \hline
    ShareGPT4V~\cite{chen2023sharegpt4v}    & GPT4-V & 0.1M & 942 \\
    ShareGPT4V-PT~\cite{chen2023sharegpt4v}   & Share-Captioner & 1.2M & 826 \\
    \cdashline{1-4}
    \textbf{ParGoCap-1M-PT}  & \textbf{GPT4-V, Gemini} & \textbf{1M} & \textbf{921} \\
    \hline
    \end{tabular}
    }}
    % \vspace{-2mm}
    \caption{
    Comparison of existing caption datasets and our ParGoCap-1M-PT. Avg. represents the average character number of the caption.
    }\label{tav:data_stat}
    % \vspace{-4mm}
\end{table}
\begin{table*}[!ht]
    \renewcommand\arraystretch{1.2}
    \centering
    \scalebox{1}{
    \begin{tabular}{l|ccc|cccccc}
        \hline
        {Method} & {LLM} & {Projector}  & {Res.} & MMB & MME$^P$ & {MME} & SEED & MM-Vet \\
        \hline
        \multicolumn{10}{l}{\textit{\textbf{MLLMs using Linear-based projectors}}} \\ 
        \hline
        LLaVA (v1)~\cite{liu2023llava}   & LLaMA-7B       & Linear  & 224  &38.7 & 502.8 & 717.5 & 33.5 & 23.8 \\
        Shikra~\cite{chen2023shikra} & Vicuna-7B       &Linear  & 224 &58.8  & - & - & - & - \\
        LLaVA-1.5~\cite{liu2023llava1.5} & Vicuna-7B  &MLP & 336 &64.3  & 1510.7 & 1795.7 & 58.3 &  30.5\\
        Honeybee~\cite{cha2023honeybee} & Vicuna-7B  & C-Abstractor & 224 &
        70.1 & \textbf{1584.2} & \underline{1891.3} & 64.5 & \textbf{34.9}\\

        \hline
        \multicolumn{10}{l}{\textit{\textbf{MLLMs using Attention-based projectors}}} \\ 
        \hline        
        MiniGPT-4~\cite{zhu2023minigpt4}  &Vicuna-7B &Resampler & 224 &24.3 & 581.7 &726.0 &47.4 & 22.1 \\
        mPLUG-Owl~\cite{ye2023mplug-owl} &LLaMA-7B &Resampler &224 &49.4 & 967.3 &1243.4 &34.0 & - \\
        InstructBLIP~\cite{dai2024instructblip} &Vicuna-7B &Q-former  &224 &36.0 &- &- &58.8 &26.2 \\
        IDEFICS &LLaMA-7B &Flamingo & 224 &48.2 &- &- &44.5 &- \\
        Qwen-VL~\cite{bai2023qwen-vl} &Qwen-7B &Resampler &448 &38.2 &- &- &62.3 &- \\
        Qwen-VL-Chat~\cite{bai2023qwen-vl} &Qwen-7B &Resampler  &448 &60.6 & 1487.5 & 1848.3 &\underline{65.4} &- \\
        Honeybee~\cite{cha2023honeybee} & Vicuna-7B & D-Abstractor  & 224 &
        \underline{70.8} &1544.1 & 1835.5 & 63.8 & - \\
        \cdashline{1-10}
        Ours & Vicuna-7B & ParGo & 336 & \textbf{73.7} & \underline{1579.86} & \textbf{1913.1} & \textbf{67.3} &  \underline{33.5} \\
        \hline
    \end{tabular}
    }
    \caption{\textbf{Comparison with other state-of-the-art MLLMs.} Res. indicates the image resolution. We highlight the \textbf{best results} and \underline{second-best results} in bold and underline.}
    % \vspace{-0.2cm}
    \label{tab:main experiments}
\end{table*}

\subsection{Implementation details}
\paragraph{Model Configuration. }
We use pre-trained EVA-02-CLIP-L/14~\cite{sun2023evaclip} with 336 resolution as the visual encoder. 
For the Large Language model, we employ the 7B Vicuna \cite{chiang2023vicuna} for a fair comparison. 
For the Partial-Global projector, six layers are utilized for all experiments if not otherwise specified. 
The number of partial and global tokens is 288 and 16, respectively, resulting in a total of 304 tokens. 

\paragraph{Training. } The proposed ParGo is trained using a two-stage pipeline, \textit{i.e.}, coarse-detailed captioned pre-training and supervised fine-tuning. 
In the pre-train stage, we freeze the visual encoder and the LLM, focusing on training the Partial-Global projector, to gain better alignment between the two modalities. 
In the supervised fine-tuning stage, the visual encoder is kept frozen, and we fine-tune the Partial-Global projector and the LLM. 
It is worth noting that instead of training the entire LLM, we use parameter-efficient fine-tuning, \textit{i.e.}, Low-Rank Adaptation(LoRA)~\cite{hu2021lora}, and the rank is set to 256. 
For all experiments, 32 A100 80GB GPUs are used. 
We employ deepspeed zero-2~\cite{rajbhandari2020deepspeedzero}  and flash-attention v2~\cite{dao2023flashattention}  for all experiments,
The pre-training stage takes approximately 24 hours, while the supervised fine-tuning tasks require around 12 hours. 
For the ablation studies,  we employ half of the training schedule (50k pre-training, 1 epoch supervised fine-tuning) compared to the final model. 
For more details, please refer to the supplementary materials.

\paragraph{Data. }  
For pre-training, we use existing coarse-captioned data, including CC-3M, SBU Caption-1M, LAION-400M, and the constructed detail-captioned data ParGoCap-1M-PT. 

For the supervised fine-tuning stage, following previous work~\cite{cha2023honeybee}, we employ four types of tasks, including:
1) Open-ended VQA, \textit{i.e.}~VQAv2~\cite{balanced_vqa_v2}, GQA~\cite{hudson2019GQA}, OCRVQA~\cite{mishra2019OCRVQA}, VSR~\cite{liu2023VSR}. 
2) Multiple-Choice VQA, including ScienceQA~\cite{lu2022ScienceQA}, A-OKVQA~\cite{schwenk2022OKVQA}. 
3) Referring Expression Comprehension (REC), which includes RefCOCO~\cite{kazemzadeh2014RefCOCO} and VG~\cite{krishna2017VG}. 
4) Instruction tuning data, LLaVA150k~\cite{liu2023llava}. 
For each dataset, we use the same templates as previous works~\cite{liu2023llava1.5, cha2023honeybee, shan2024mctbench}.

\subsection{Main Results}
In Table~\ref{tab:main experiments}, we compare our ParGo with previous state-of-the-art methods. 
Firstly, we compare our model with the methods that use the attention-based projector. 
Compared with Honeybee (D-Abstractor), our ParGo obtains significant improvement, \textit{e.g.}, $2.9$ in MMBench and $77.6$ in MME. 
Furthermore, compared with methods that use vanilla attention-based projector \textit{e.g.,} Resampler or Q-former, our advantages are more apparent, \textit{e.g.,} $37.7$ in MMBench compared with InstructBLIP~\cite{dai2024instructblip}. 

Additionally, we compare our model with the methods that use linear-based projectors, as shown in Table~\ref{tab:main experiments} top. 
Compared with methods using Linear or MLP projectors, the advantage of our Partial-Global projector is also significant, \textit{i.e.} 117.4 improvement in MME compared with LLaVA-1.5\cite{liu2023llava1.5}. 
Compared with a more recent work, Honeybee (C-Abstractor)~\cite{cha2023honeybee}, our ParGo outperforms in MMB, MME, and SEED, while slightly underperforming in MM-Vet.
It is worth noting that honeybee~\cite{cha2023honeybee} fine-tuning all the LLM parameters during the supervised fine-tuning state, while we use the parameter efficient fine-tuning strategy, \textit{i.e.,} Low-Rank Adaptation~(LoRA)~\cite{hu2021lora}. 
This full-finetuning strategy boosts the performance of Honeybee in the MM-Vet benchmark, which needs more natural language generation ability.  

These consistent results demonstrate the superiority of our proposed Partial-Global projector in effectively bridging the representation gap between visual and language modalities. 

\subsection{Ablation Study}
\paragraph{Effect of the components in Partial-Global projector. }
In Table~\ref{tab:Global_Partial_Visual Projector}, we ablate the main component in our proposed Partial-Global projector, \textit{i.e.}, the Partial-Global Perception (PGP) block and Cascaded Partial Perception (CPP) block. 
For a fair comparison, the baseline uses a Q-Former with 304 tokens as the visual projector. 
As shown in the table, by adding the Partial-Global Perception block (\textit{i.e.,} 16 global tokens and 288 partial tokens), we obtain a significant improvement over the baseline, \textit{i.e.,} 162.73 in MME. 
These results suggest that our proposed Partial-Global projection effectively bridges the representation gap, boosting the overall performance. 
Then, by introducing the Cascaded-Partial Perception block, our model further obtains an improvement, \textit{i.e.,} 92.23 in MME. 
In summary, the ablation study demonstrates
the effectiveness of the proposed Partial-Global projector.

\begin{table}[!t]
  \renewcommand\arraystretch{1.2}
  \centering
  \begin{tabular}{cc|cc}
    \hline
     PGP & CPP & MME & MM-Vet \\
    \hline
       &  & 1591.74 & 28.3 \\
    \checkmark     &  & 1754.47 & 32.7   \\
    \checkmark     & \checkmark   &  1851.70  &  33.1      \\
    \hline
  \end{tabular}
  \caption{Ablations on the Partial-Global Perception (PGP) block and the Cascaded Partial Perception (CPP) block. The baseline uses a Q-Former as the projector.} 
  \label{tab:Global_Partial_Visual Projector}
  % \vspace{-0.3cm} 
\end{table}

\begin{table}[!t]
  \renewcommand\arraystretch{1.1}
  \centering
  \begin{tabular}{cc|c|cc}
    \hline
    \multicolumn{3}{c|}{Number of Tokens} & \multirow{2}{*}{MME} & \multirow{2}{*}{MM-Vet}\\
    \cline{0-2}
    Global & Partial & Total  & \\
    \hline
    160    &  -    &  160 &  1561.42  &  30.1     \\
     -     & 144   &  144 &  1681.55  &  30.5     \\
    16     & 144   &  160 &  1802.37  &  32.8      \\
    16     & 288   &  304 &  1851.70  &  33.1  \\
    \hline
  \end{tabular}
  \caption{Ablations on the number of partial and global tokens. 
  Total indicates the sum of partial and global tokens.}
  \label{tab:Global_Local_Tokens}
  % \vspace{-0.4cm}
\end{table}

\paragraph{Effect of the number of Global Partial tokens. }
In Table~\ref{tab:Global_Local_Tokens}, we conduct a quantitative analysis of the number of partial and global tokens. 
Firstly, for a fair comparison, we conduct an experiment using only 160 global tokens, which achieve 1561.42 scores in MME. 
Then, by replacing 144 global tokens with 144 partial tokens (still 160 tokens in total), the model achieves a significant improvement (\textit{i.e.,} 140.95 score in MME), which demonstrates the effectiveness of our partial tokens in preserving partial information. 
Finally, we scale up the number of partial tokens, using 288 partial tokens and 16 global tokens, and the model achieves a further improvement. 
Additionally, we conduct an experiment that uses only 144 partial tokens, which still outperforms the baseline with 160 global tokens, \textit{i.e.,} 1681.55 \textit{vs.} 1561.42 in MME. 
These consistent results demonstrate the superiority of our proposed partial tokens in preserving the partial information. 

\paragraph{Effect of the pre-training data. }
In our method, to facilitate the simultaneous learning of the partial and global information, we use two kinds of pre-training data, coarse-captioned and detail-captioned.
In Table~\ref{tab:pretrain_data}, we ablate the pre-training data to verify the effectiveness of this strategy. 
As shown in the table, by adding the detail-captioned data, the model achieves an improvement in both benchmarks, \textit{i.e.,} 116.84 in MME and 0.9 in MM-Vet.
It is worth noting that the training configurations (\textit{e.g.,} batch size and training steps) used for these experiments are the same, ensuring that the models are trained with the same amount of data. 
These results demonstrate the detail-captioned data help the model align the visual and language modalities. 

\begin{table}[!t]
  \renewcommand\arraystretch{1.2}

  \centering
  \begin{tabular}{cc|cc}
    \hline
    \multicolumn{2}{c|}{Pre-training Data} & \multirow{2}{*}{MME} & \multirow{2}{*}{MM-Vet}\\
    \cline{0-1}
    Coarse & Detailed &  &  \\
    \hline
         % &  \checkmark  & - \\
    \checkmark     &     & 1734.86  & 32.2     \\
    \checkmark     & \checkmark   & 1851.70 &  33.1  \\
    \hline
  \end{tabular}
  \caption{Ablations on the pre-training data. We use different combinations of Coarse captioned data and Detailed captioned data. }
  \label{tab:pretrain_data}
  % \vspace{-0.3cm}
\end{table}

\subsection{Analysis}
\begin{table}[!t]
  \renewcommand\arraystretch{1.2}
  \centering
  % \scalebox{0.95}{
  \begin{tabular}{c|c|cc}
    \hline
    Projectors & \# Tokens  & MME & MM-Vet \\
    \hline
    Linear            & 576   &  1727.83 & 32.8   \\
    Q-Former          & 304   &  1591.74 & 28.3   \\
    ParGo             & 304   &  1851.70 & 33.1   \\
    \hline
  \end{tabular}
  % }
  \caption{Comparison of our Partial-Global projector (ParGo) with existing Linear and Q-former projector. 
  \# Tokens means the number of tokens the projector outputs.}
  \label{tab:linear_qformer_ours}
  % \vspace{-0.4cm}
\end{table}

\paragraph{Comparison with existing visual projectors. }
To further illustrate the effectiveness of our proposed Partial-Global projector, in Table~\ref{tab:linear_qformer_ours}, we compare our proposed projector with a linear projector and a Q-former projector, while using the same training data and schedule for a fair comparison. 
The linear projector is a simple linear layer that produces 576 tokens; the Q-Former is 6-layer and produces 304 tokens. 
As shown in the table, the linear-based projector performs much better than the Q-Former as it uses a one-to-one projection that directly project the visual feature to the language feature space, with less information loss. 
However, the one-to-one projection also results in a huge number of visual tokens being fed into the LLM, introducing large computation costs. 
Compared with our proposed ParGo, the linear-based underperforms in both benchmarks. 
Our Pargo achieves an increase of 259.96 points in MME compared with the attention-based Q-Former projector. 
These results highlight the effectiveness of our ParGo in preserving visual information while utilizing fewer visual tokens, demonstrating its superior capability in aligning the vision and language modalities.

\paragraph{Comparison of different base Large Language Models.}
Here, we verify the generalizability of our proposed projector, we ablate several widely-used Large Language Models (LLMs), \textit{i.e.}, Vicuna-7B~\cite{chiang2023vicuna}, Llama3-8B~\cite{dubey2024llama3}, internLM2-7B~\cite{cai2024internlm2}. 
As shown in Table~\ref{tab:llm}, with stronger LLM, the overall performance is further improved. 
Employing internLM2-7B as the base LLM achieves the best performance in both benchmarks, \textit{i.e.,} 1869.76 in MME and 37.0 in MM-Vet. 
These results demonstrate the generalizability of the Partial-Global projector in aligning the visual and language modalities. 

\begin{table}[!t]
  \renewcommand\arraystretch{1.2}

  \centering
  \begin{tabular}{c|cc}
    \hline
    Large Language Model & MME & MM-Vet\\
    \hline
    Vicuna-7B~\cite{chiang2023vicuna} & 1851.70  & 33.1 \\
    Llama3-8B~\cite{dubey2024llama3}  & 1866.43 & 35.9\\
    InternLM2-7B~\cite{cai2024internlm2}& 1869.76  & 37.0     \\
    \hline
  \end{tabular}
  \caption{Comparison of different Large Language Models.}
  \label{tab:llm}
  % \vspace{-0.2cm}
\end{table}

\begin{table}[!t]
  \renewcommand\arraystretch{1.2}
  \centering
  % \scalebox{0.95}{
  \begin{tabular}{c|cccc}
    \hline
    \multirow{2}{*}{Projectors}  & \multicolumn{3}{c}{MME} & \multicolumn{1}{c}{MM-Vet} \\
    &  CNT & OCR & EXST & REC\\
    \hline
    Linear  & 135.0 & 137.5 & 190 & 37.2 \\
    Q-Former     & 128.33 & 130.0 & 175    & 34.4     \\
    ParGo   & 146.66 & 162.5 & 190  & 37.6 \\
    \hline
  \end{tabular}
  % }
  \caption{Analysis on the detail perception ability. 
  We select 4 tasks that require detail perception ability from the MME and MM-Vet benchmark, including Count (CNT), Optical Character Recognition(OCR), and Existence (EXST) from MME, Recognition (REC) from MM-Vet Benchmark.}
  \label{tab:Quantitative_Paritial}
  % \vspace{-0.3cm}
\end{table}

\paragraph{Analysis on the Partial-Global Projector.} 
To thoroughly examine the superiority of our proposed Partial-Global projector, in Table~\ref{tab:Quantitative_Paritial}, we assess the ability of different projectors in image detail perception by selecting 4 tasks that require more detail perception ability from the MME and the MM-Vet benchmarks. 
As shown in the table, in the tasks of OCR and CNT, which require the most detailed perception, our ParGo achieves a substantial improvement compared to the Linear projector and Q-Former Projector, \textit{i.e.,} 16.33 in CNT and 32.5 in OCR compared with Q-Former. 
In the other two tasks, EXST and REC, our model also achieved better performance compared to other projectors.
These results demonstrate the effectiveness of the partial-global projection of our ParGo, which enhances the perception of image details while maintaining global perception. This dual capability leads to superior performance across diverse tasks.

\begin{figure*}[!t]
  \centering
  \includegraphics[width=0.8\linewidth]{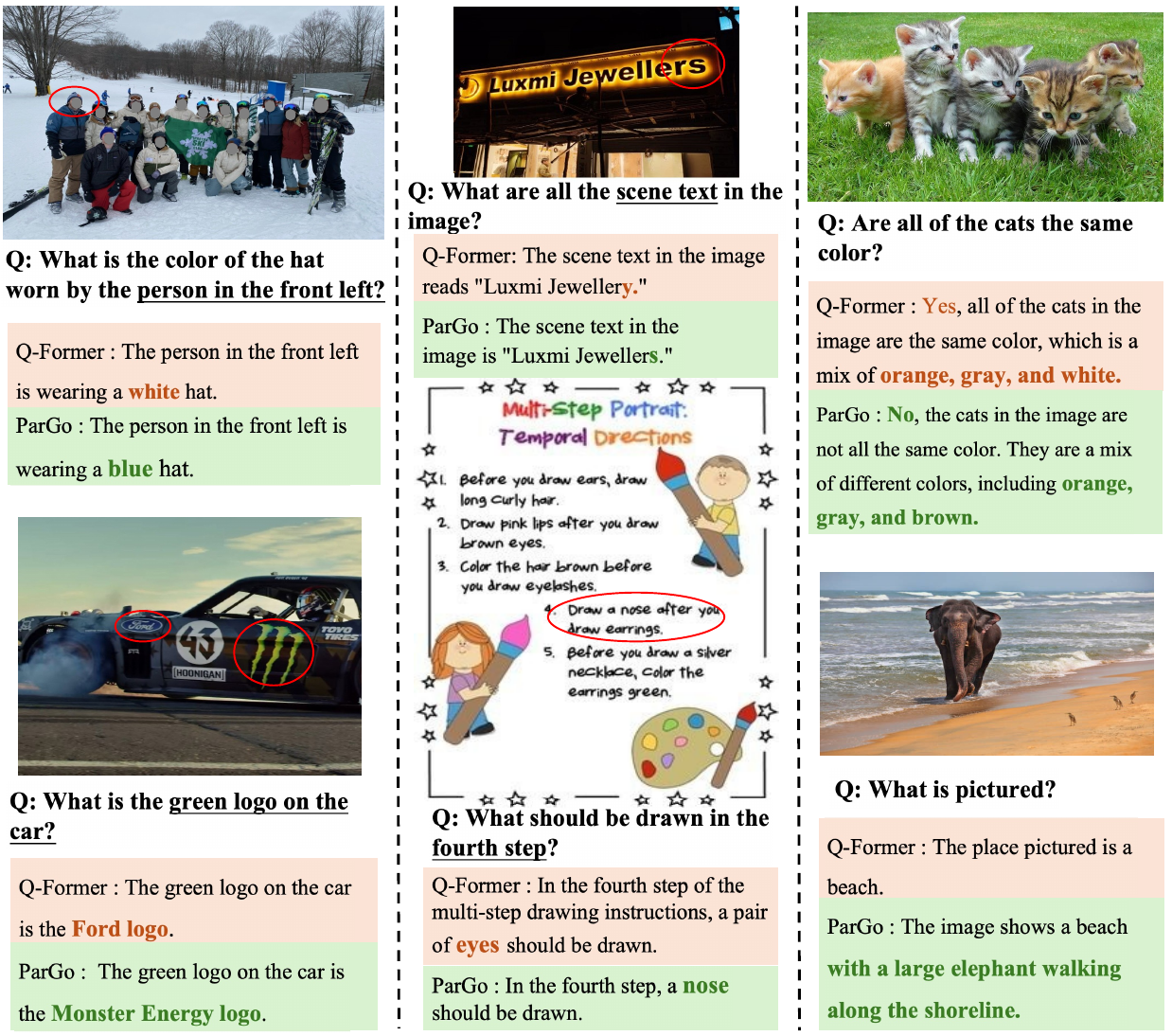}
  \caption{
  Case study on the proposed Partial-Global projector (ParGo). 
  In this figure, we select 6 examples to illustrate the superiority of our proposed ParGo in aligning vision and language modalities. 
  }
  \label{fig:Qualitative}
  \vspace{-0.2cm}
\end{figure*}

\subsection{Case Study}
In Figure~\ref{fig:Qualitative}, we give several examples to demonstrate the superiority of our proposed Partial-Global projector in preserving partial and global information. 
Specifically, we compare our full model with the baseline that uses a Q-Former with 304 tokens as the projector. 
From the first two examples (in the first column), we find that our model has a better perception of the location-specified image part (\textit{i.e.}, the person in the front left, the green logo on the car.).
Conversely, due to the Q-Former projector's overemphasis on prominent regions and neglect of image details, it fails to answer the question correctly. 
Regarding the two examples in the middle column, our model accurately identifies the characters in the image and correctly understands the text. 
In contrast, the Q-Former fails to achieve correct recognition in these instances.
The last two examples illustrate that our method has a better perception of the overall images while maintaining the partial information, \textit{i.e.,} the colors of different cats and the presence of the elephant on the shoreline.
In summary, the qualitative results highlight the ability of our proposed Projector to align the two modalities, providing the LLM with features covering both partial and global information.

\section{Conclusion}
In this work, we focus on the vision-language projector in MLLMs, proposing Partial-Global projector (ParGo). 
ParGo employs partial and global tokens with specially designed attention masks to extract two kinds of information separately, with considering the relation between different partial regions in an image. 
Moreover, to further facilitate the alignment between the two modalities, we contribute a large-scale detail-captioned dataset ParGoCap-1M-PT for pre-training. 
Extensive ablations and experiments are conducted, which illustrate the effectiveness of our ParGo. 
% For instance, our ParGo outperforms Q-Former by 259.96 scores in the MME benchmark. 
We find that ParGo significantly outperforms other projectors, particularly in tasks that emphasize detail perception. 
These results highlight ParGo's potential to enhance MLLMs by providing a more nuanced understanding of visual content through the integration of both partial and global views.
\section{Acknowledgments}
This work was partially supported by the National Key Research and Development Program of China (2023YFA1008503), NSFC(92470202, U21A20471), and Guangdong NSF Project (No. 2023B1515040025).
\bibliography{main}

\end{document}